\documentclass[letterpaper]{article}
\usepackage{times}
\usepackage{helvet}
\usepackage{courier}
\usepackage{adjustbox}
\usepackage{geometry}
\usepackage{array}
\usepackage{booktabs}
\usepackage{graphicx}
\usepackage{caption}
\usepackage[table,xcdraw]{xcolor}
\usepackage{amssymb}
\usepackage{threeparttable}
\usepackage{hyperref}
\usepackage{placeins}
\usepackage{amsmath}
\usepackage{aaai}

\hypersetup{
  colorlinks=true,   
  linkcolor=black,   
  citecolor=black,   
  urlcolor=blue      
}

\begin{document}
%
\title{Skin Cancer Detection utilizing Deep Learning: Classification of Skin Lesion Images using a Vision Transformer}
\author{Carolin Flosdorf, Justin Engelker, Igor Keller, Nicolas Mohr\\
University of Cologne\\
Albertus-Magnus-Platz\\
50923 Cologne\\
ikeller@smail.uni-koeln.de\\
}
\maketitle
\begin{abstract}
\begin{quote}
Skin cancer detection still represents a major challenge in healthcare. Common detection methods can be lengthy and require human assistance which falls short in many countries. Previous research demonstrates how convolutional neural networks (CNNs) can help effectively through both automation and an accuracy that is comparable to the human level. However, despite the progress in previous decades, the precision is still limited, leading to substantial misclassifications that have a serious impact on people’s health. Hence, we employ a Vision Transformer (ViT) that has been developed in recent years based on the idea of a self-attention mechanism, specifically two configurations of a pre-trained ViT. We generally find superior metrics for classifying skin lesions after comparing them to base models such as decision tree classifier and k-nearest neighbor (KNN) classifier, as well as to CNNs and less complex ViTs. In particular, we attach greater importance to the performance around melanoma, which is the most lethal type of skin cancer. The ViT\_L32 model achieves an accuracy of  91.57\% and a melanoma recall of 58.54\%, while ViT\_L16 achieves an accuracy of  92.79\% and a melanoma recall of 56.10\%. This offers a potential tool for faster and more accurate diagnoses and an overall improvement for the health care sector.

\end{quote}
\end{abstract}

\section{Introduction}
Skin cancer is the most common group of cancer diagnosed worldwide, with over 1.5 million new patients in 2020 alone \cite{IARC_SkinCancer}. Since the risks of the disease vary significantly by type, it is essential to identify the correct kind of skin cancer early. Once this information is available, proper treatment can be ensured. Generally, there are three broad types including basal cell carcinoma (BCC), squamous cell carcinoma (SCC) and melanoma. While most people die from the most common types BCC and SCC, melanoma poses a much greater threat with over 60,000 deaths out of 325,000 cases globally each year. In addition, the detection process is often lengthy as it requires an appointment and the time of a doctor. Automated diagnoses may speed up this process and address increasing shortages the healthcare system is regionally suffering from, thus potentially saving more lives \cite{ostwald2010fachkraftemangel}.

Accurately classifying various types of skin cancers, including melanoma, is a challenging task that requires nuanced differentiation from benign lesions. The variability in skin lesions, influenced by factors such as skin type, lesion location, and individual patient characteristics, adds to the complexity. Additionally, the sheer volume of cases globally creates a pressing need for scalable and efficient diagnostic methods. Previous studies mainly concentrates on convolutional neural networks (CNN). Although these networks significantly improve the accuracy of diagnosis compared to manual diagnosis, their limitations in terms of accuracy and specificity, particularly for melanoma, remain a concern as these metrics still hold potential for improvement. \citeauthor{haenssle2018man} \shortcite{haenssle2018man} and \citeauthor{garg2021decision} \shortcite{garg2021decision} demonstrate the potential of CNNs for improving diagnostic precision in their publications. However, more refined approaches are needed to achieve this goal.

Therefore, we join the mission of accurately detecting skin cancer and its subtypes by employing deep learning. The Skin Cancer MNIST: HAM10000 dataset is commonly used for the classification of skin cancer images, which we also utilize in our project \cite{codella2019skin,tschandl2018ham10000}. It contains approximately 10,000 images of 7 different skin cancer types. In our approach, we enrich the number of images with the help of data augmentation. While many projects rely on CNNs, we contribute to the existing literature by using an approach that has received less attention. Specifically, we build models based on two larger configurations of a pre-trained Vision Transformer (ViT), which are trained on several million images with tens of thousands of classes. This deep learning architecture for computer vision tasks features self-attention mechanisms that have gained popularity in recent years. ViT processes images as sequences of patches and handles them like tokens in natural language processing. This can lead to the superiority of these models in image classification compared to previous approaches \cite{dosovitskiy2020image}. For the smallest configurations of a pre-trained ViT, an application already exists, which we use together with a CNN-based model to evaluate our models.

The implications of the findings are significant. If validated in clinical settings, ViT models could offer a fast, accurate, and accessible tool for skin cancer screening, benefiting healthcare systems worldwide, especially in regions with limited access to dermatologists. This innovation holds promise not just for skin cancer but as a model for applying advanced machine learning techniques to other diagnostic challenges in medicine.

\section{Related Work}
There are a variety of machine learning approaches in research that aim to classify images with diseases caused by skin lesions in general. Several publications use clinical images for a CNN approach \cite{brinker2018skin,han2018classification}. The successful performance of CNN-based models is also confirmed by their superior performance in skin cancer classification compared against 58 international dermatologists \cite{haenssle2018man}. 

For the classification of images from the Skin Cancer MNIST: HAM10000 dataset in particular, there are similarly neural network-driven approaches \cite{codella2019skin,tschandl2018ham10000}. As of December 2023, the Kaggle page of the dataset contains over 475 codes \cite{mader2020skin}. In research, this dataset is also used frequently in combination with a CNN approach to classify the skin cancer images \cite{shete2021detection,nugroho2019skins,huo2021full}. \citeauthor{garg2021decision} \shortcite{garg2021decision} achieve a highly successful result with their CNN-based approach in combination with augmentation strategies to increase the number of images and also use transfer learning methods such as Residual Neural Network (ResNet).

In addition, the newer approach from 2020 for image classification called ViT is becoming increasingly relevant in the field. Empirical evidence shows that this approach leads to better results in classification tasks compared to common methods such as CNN \cite{dosovitskiy2020image}.
\citeauthor{korgialas2021vision} \shortcite{korgialas2021vision} uses in his work the configurations ViT\_B16 and ViT\_B32 as well as an augmentation approach to assign the images from the Skin Cancer MNIST: HAM10000 dataset to one of the seven skin cancer types. 

Overall, primarily existing research mainly uses the established CNN approach for classification or the smaller pre-trained ViT configurations. Furthermore, the existing work does not explicitly assess the performance of the respective model for the relatively most lethal skin cancer type of the dataset, melanoma.

\section{Methodology}

\begin{figure}[htbp]
    \centering
    \includegraphics[width=0.45\textwidth]{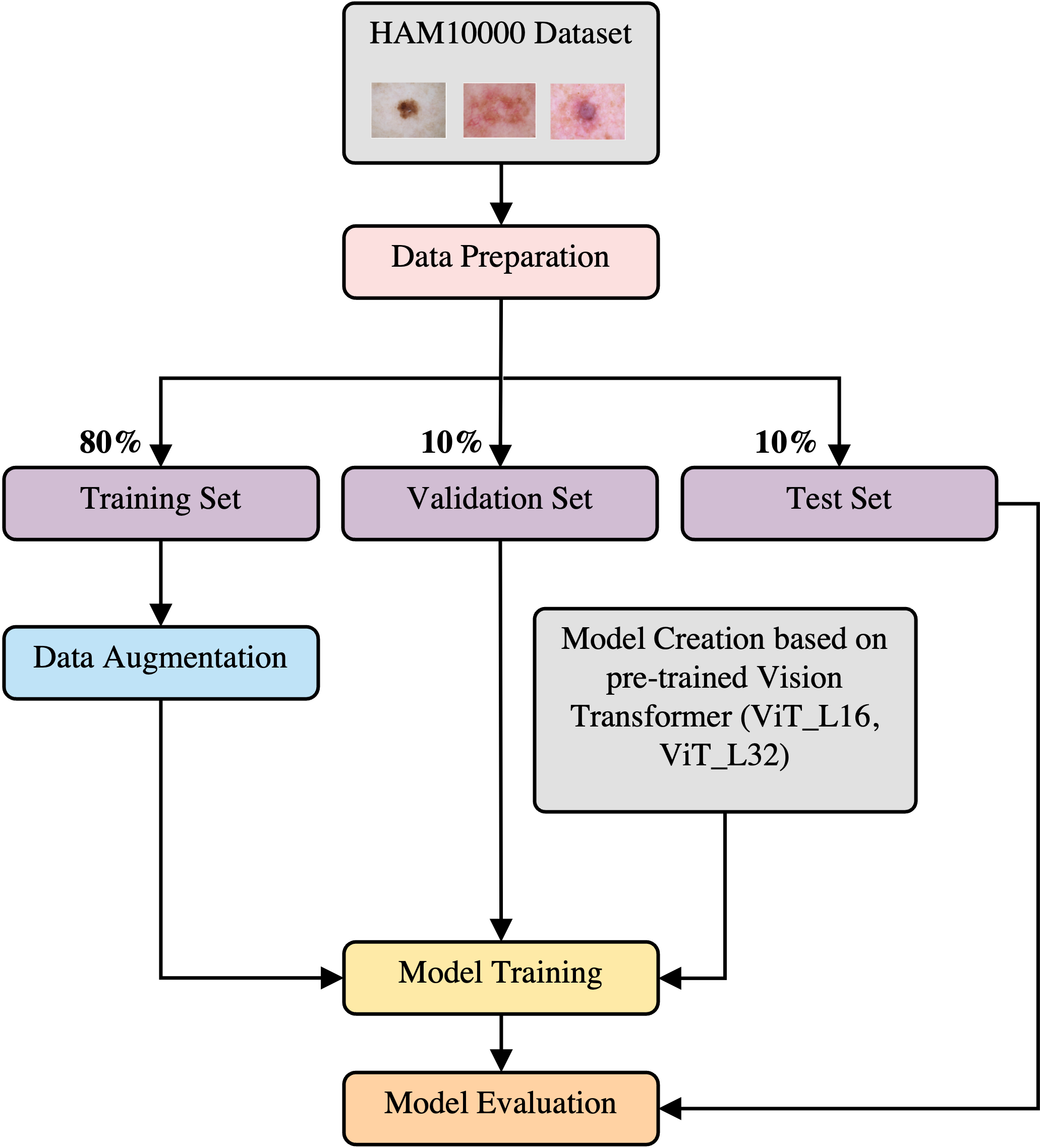}
    \caption{Pipeline}
    \label{fig:1}
\end{figure}

Our approach, as can be seen in Figure \ref{fig:1}, comprises different steps, which are presented in more detail in the following sections. We start with data preparation, which includes data cleansing. The dataset is then split into different subsets, with 80\% used for the training set and 10\% each for the validation and test set. The training dataset is then further processed by data augmentation. The next step is model generation, where we use pre-trained ViT models, in particular ViT\_L16 and ViT\_L32 architectures. In the training phase, the model parameters are optimized using the prepared training set, while the performance of the model is observed using the validation dataset. Various techniques are used in this phase to ensure efficient training. Finally, the model is evaluated on the test dataset. The following section provides further details on these steps.

\begin{figure}[htbp]
    \centering
    \includegraphics[width=0.45\textwidth]{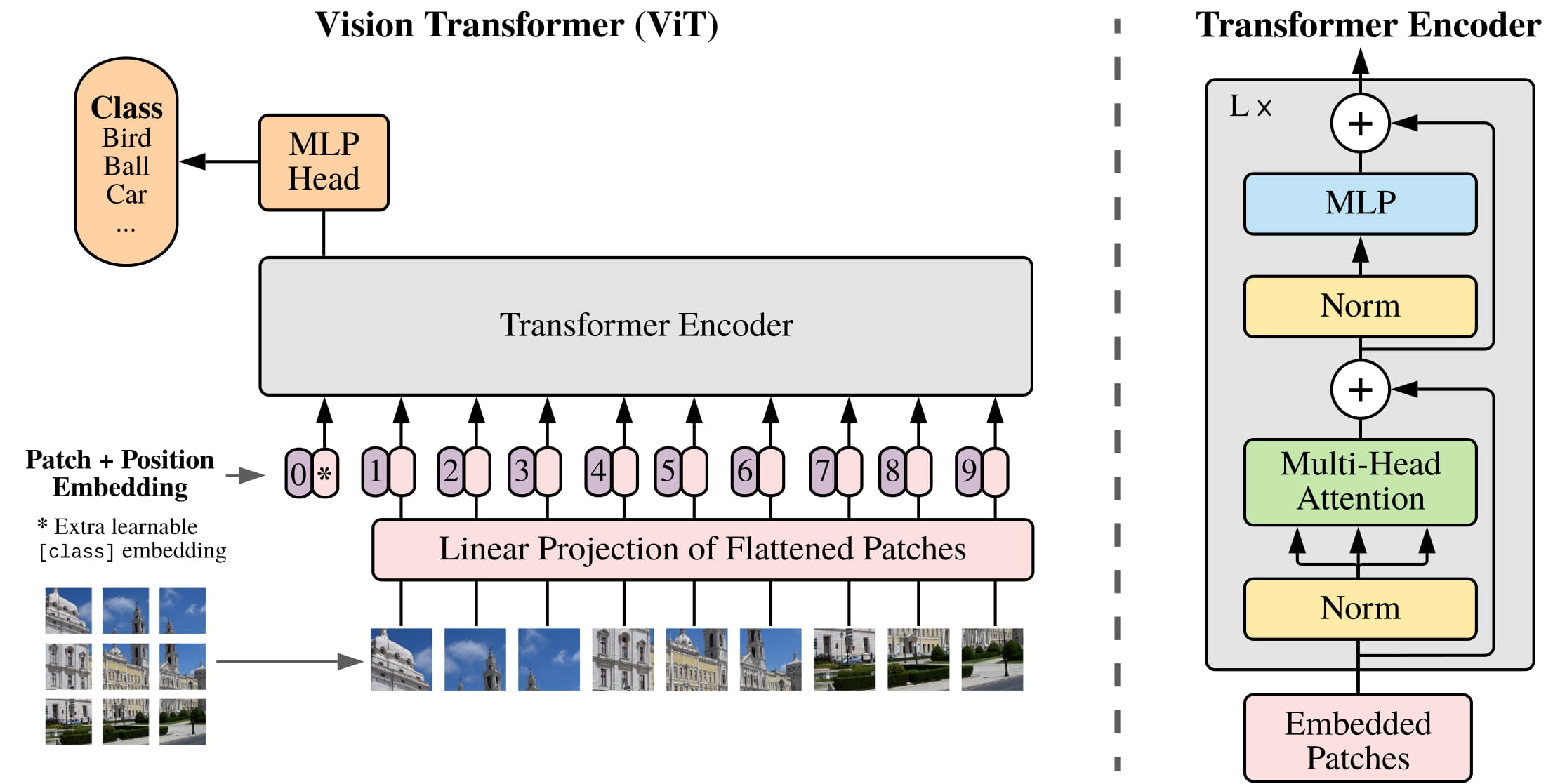}
    \caption{Vision Transformer by \citeauthor{dosovitskiy2020image} \shortcite{dosovitskiy2020image} and Transformer Encoder by \citeauthor{vaswani2017attention} \shortcite{vaswani2017attention}}
    \label{fig:2}
\end{figure}

This machine learning image classification project uses a ViT as its fundamental method. The general functionality of a ViT is presented for the first time in a paper by \citeauthor{dosovitskiy2020image} \shortcite{dosovitskiy2020image} and can be seen in Figure \ref{fig:2}. At the beginning, an image is divided into smaller fixed-sized areas (patches). Each of the patches represents a region of the image. The pixel values within each patch are then reduced to a single vector. This allows the image patches to be treated as sequential data. A lower dimensional linear embedding is then produced from the flattened patches. This reduces the dimensionality of the data while retaining the important features. Position embeddings are subsequently added. This provides information about the spatial arrangement of the patches, which helps the model to understand the relative positions of the different patches in an image. The sequence of patch embeddings and positional embeddings is then entered into a standard transformer encoder as published by \citeauthor{vaswani2017attention} \shortcite{vaswani2017attention}. This encoder is composed of several layers, which contain two important components. Firstly, the multi-head self-attention mechanisms (MSPs). This is responsible for calculating attention weights so that input sequence elements are prioritized during prediction. And secondly, the multi-layer perceptron (MLP) blocks. To ensure stability and efficiency during training, a layer normalization (LN) is applied before each MLP block to scale and center the data appropriately. An optimizer is also used to adjust the hyperparameters of the model during training. The output from the Transformer encoder is not sent to a decoder, but to an MLP-head. This is added to the model sequence and serves as an additional adaptive classification token to work as a classifier.

The two models included in this project are created in the Python environment and use the configurations ViT\_L16 and ViT\_L32 of the pre-trained ViT models from the open source deep learning library Keras (Chollet et al., 2015). These differ from the ViT\_B16 and ViT\_B32 configurations in having 1024 hidden dimensions instead of 768, as well as 4096 instead of 3072 MLP dimensions. In addition, they have 16 attention heads, i.e. 4 more, and the encoder depth is twice as large at 24. So, the two configurations used in this project result in larger models. ViT\_L16 and ViT\_L32 differ in that the patch size is 16×16 for the former and 32×32 for the latter \cite{chollet2015keras}. The additional structure of both variants is identical. At the start of the model creation, a distinction is made between the two configurations. As both pursue the same goal of assigning the images to one of the seven skin cancer types, the class's argument is set to '7'. The size of the images is 224×224, which means that patch size can be evenly divided by this. This is a recommended specification \cite{chollet2015keras}. The activation function is set to softmax. The respective models are initialized with pre-trained weights, these are initially trained on the basis of \textit{imagenet21K}, which has over 14 million images with 21,843 classes and is fine-tuned with the help of \textit{imagenet2012}, which has 1 million images with 1,000 classes \cite{chollet2015keras}. In addition, we specify in the function that the basic framework of the model, i.e. all layers except the last classification layer, is loaded with pre-trained weights and that the last classification layer is omitted. This is useful so that we can add our own user-defined classification layer. These two basic ViT models described serve as the first part of a sequential extension. The respective ViT model is followed by a flatten layer, which lowers the input from a 3D tensor to a 1D tensor. The flatten layer is followed by a batch normalization layer to normalize and stabilize the activations of the network. It is further extended by a dense layer (fully connected) with 28 neurons. The scientifically established non-linear rectified linear unit (ReLU) is applied as the activation function \cite{james2013introduction}. Its definition is as follows:

\begin{equation}
ReLU(z) = \max(0, z)
\end{equation}

Again, a batch normalization layer is added, followed by a dropout layer and the final dense layer that has 7 (skin cancer types) neurons and the activation function softmax to facilitate the interpretation of the output values as class probabilities, as this is a crucial aspect in multi-class classification tasks \cite{goodfellow2016deep}. The softmax function for this problem is designed according to the following equation: 

\begin{equation}
\sigma(z_i) = \frac{e^{z_{i}}}{\sum_{j=1}^7 e^{z_{j}}} \quad \text{for } i=1,2,\dots,7
\end{equation}

When compiling the respective model, the optimizer is stochastic gradient descent (SGD). Since this is a multi-class classification problem, we use categorical cross entropy as the loss \cite{ho2020real}. Three callbacks are integrated for the training of the respective neural network. The first is the early stopping, which stops the training if the validation loss during training does not improve for five consecutive epochs. This has the great advantage that it helps with overfitting, especially when using data augmentation \cite{rice2020overfitting}. Also included is an option to save the weights of the model whenever they contribute to an improvement in validation accuracy. The third callback modifies the learning rate if it is on a plateau during training and the validation loss does not improve by three consecutive epochs. Finally, the number of epochs for training is set to 20 and the number of batches to be processed in a training and validation epoch is 16 in both cases. 

The performance evaluation focuses on the accuracy metric, measuring the proportion of correctly classified observations by the model. The accuracy for the respective trained model is calculated using the test data and is compared to a decision tree classifier (DTC) and a k-nearest-neighbor (KNN) classifier, which serve as base models. Due to the high number of CNN approaches for this problem, we compare our results only to the paper by \citeauthor{garg2021decision} \shortcite{garg2021decision}, which contains the highest performing approach among the considered works. In addition, the results of the ViT with the smaller configurations of \citeauthor{korgialas2021vision} \shortcite{korgialas2021vision} are also used for comparison. Here, however, it should be noted that we use different data for the test and validation set; this does not apply to the project by \citeauthor{korgialas2021vision} \shortcite{korgialas2021vision}. By doing so, we increase the reliability of performance across different samples \cite{stimpfl1995validation}. Since our project, in contrast to the works we use for comparison, also focuses on how the best accuracy models perform for the relatively deadliest form of skin cancer, melanoma, we use in a second evaluation the metric recall (sensitivity) to assess this. We evaluate the recall of melanoma predictions by comparing the number of true positive cases to the total number of actual cases, which includes both true positive and false negative cases (another skin cancer type is predicted but it is in fact melanoma) \cite{james2013introduction}. We believe this is a crucial addition because a misdiagnosis that fails to identify melanoma when it is present carries more weight than a diagnosis that identifies melanoma when it is not present. This is because the first variant harbors a significantly higher mortality risk.

\section{Experiment}

\begin{figure}[htbp]
    \centering
    \includegraphics[width=0.45\textwidth]{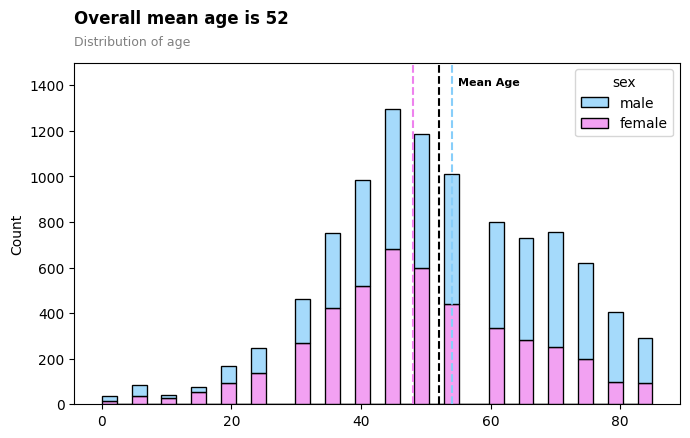}
    \caption{Age Distribution}
    \label{fig:3}
\end{figure}

Before delving directly into the main analysis, we provide an exploratory data analysis (EDA) to get an understanding of the data used. The Skin Cancer MNIST: HAM10000 dataset (source) serves as the basis for our analysis and comprises a large (N=10,015) collection of dermatoscopic images as well as additional demographic features of patients \cite{codella2019skin,tschandl2018ham10000}. This dataset is accessed directly from Kaggle using an application programming interface \cite{mader2020skin}.
After removing missing values (e.g., unknown gender), we end up with 9,948 observations of which 54\% are male (Appendix A: Figure \ref{fig:5}). Figure \ref{fig:3} illustrates the age distribution which looks approximately normal with the male group being around five years older on average and an overall mean of 52 which makes sense since people at that age are more likely to suffer from skin cancer. However, the most striking aspect in Figure \ref{fig:3} is the wide age range emphasizing the importance for society as a whole. Not only adults and older people but also children and babies are affected, even though in fewer cases.

In total, we have seven labels for our response variable indicating the type of skin lesion. We clearly see that most instances are benign (Appendix A: Figure \ref{fig:6}) while males and females appear to be equally affected by each type (Appendix A: Figure \ref{fig:7}). In contrast to the population where BCC is the most prevalent type, it is underrepresented in the data whereas melanoma is far overrepresented. Hence, applying a model that has been trained on these data may lead to errors on new unseen data that is more similar to its population. On the other hand, the data confirms our expectations that the likelihood of malignant skin cancer increases with age (Appendix A: Figure \ref{fig:8}). Lastly, the most common localization area is the back, especially for men (Appendix A: Figure \ref{fig:9}; Appendix A: Figure \ref{fig:10}) followed by lower extremity and trunk. These areas on the body occur quite consistently across gender and age groups (Appendix A: \ref{fig:10}; Appendix A: \ref{fig:11}).

Once the dataset has been cleansed and the EDA created, the unique images are filtered. This procedure prevents identical images from being in different sets. Afterwards, during the data split, 985 observations are assigned to the test set (10\% of the total data), 987 are allocated to the validation set (10\%) and 7976 to the training set (80\%) prior to the application of data augmentation. This is in line with common practice \cite{baheti2022train}. We use data augmentation to address the uneven distribution of skin cancer types in our training dataset. We create additional images for the six lesser-represented classes, aside from melanocytic nevi. This strategy enriches the dataset and potentially minimizes overfitting risks with the original data. However, we acknowledge possible overfitting concerns for the augmented data \cite{he2019data}. The augmentation techniques that are used in this study include random rotations of up to 180 degrees, horizontal and vertical shifts, rotation and shearing transformations, random brightness adjustments, and zoom. Additionally, boundary outliers are filled using nearest points. These augmentation techniques are employed to ensure a more balanced class distribution, mitigating the dominance of melanocytic nevi.

The relevant evaluation metric for the models is accuracy, which are shown in Table \ref{tab:models_comparison} for the respective models. The calculation of this metric involves looking at the number of correct predictions, i.e. true positives (TP) and true negatives (TN), in relation to the total number of predictions, i.e. true positives (TP), true negatives (TN), false positives (FP) and false negatives (FN) and is in mathematical terms defined as follows:

\begin{equation}
Accuracy = \frac{TP + TN}{TP + TN + FP + FN}
\end{equation}

In addition, the recall for cancer class melanoma is also shown in Table \ref{tab:models_comparison}. This is calculated according to the following formula:

\begin{equation}
Recall = \frac{TP}{TP + FN}
\end{equation}

\begin{table}[htbp]
    \centering
    \caption{Comparison of Different Models}
    \label{tab:models_comparison}
    \scriptsize 
    \begin{adjustbox}{width=\columnwidth}
    \begin{tabular}{lccccccc}
        \toprule
        Model & DTC & KNN-Classifier & \textbf{ViT\_L32} & \textbf{ViT\_L16} & ViT\_B32* & ViT\_B16* & CNN** \\
        \midrule
        Accuracy & 61.06\% & 65.45\% & \textbf{91.57\%} & \textbf{92.79\%} & 74.73\% & 81.88\% & 90.51\% \\
        \midrule
        Recall & 24.78\% & 6.19\% & \textbf{58.54\%} & \textbf{56.10\%} & 41.03\% & 17.95\% & 57.57\% \\
        \bottomrule
    \end{tabular}
    \end{adjustbox}
    \begin{threeparttable}
        \scriptsize
        \begin{tablenotes}
            \item[*] Results from \citeauthor{korgialas2021vision} \shortcite{korgialas2021vision}. These do not determine the recall metric for melanoma. We compute it manually. Different data for test and validation sets.
            \item[**] Results from \citeauthor{garg2021decision} \shortcite{garg2021decision}. These do not determine the recall metric for melanoma. We compute it manually.
        \end{tablenotes}
    \end{threeparttable}
\end{table}

The results show that the base models, DTC and KNN classifier, cannot easily solve the underlying problem in an optimal way. The respective accuracy is 61.06\% for the former and 65.45\% for the latter base model. The two focused models in this report, which are based on a pre-trained ViT, perform significantly better. The ViT\_L32 model achieves an accuracy of 91.57\% and ViT\_L16 an accuracy of 92.79\%. 

\begin{figure}[htbp]
    \centering
    \includegraphics[width=0.8\columnwidth]{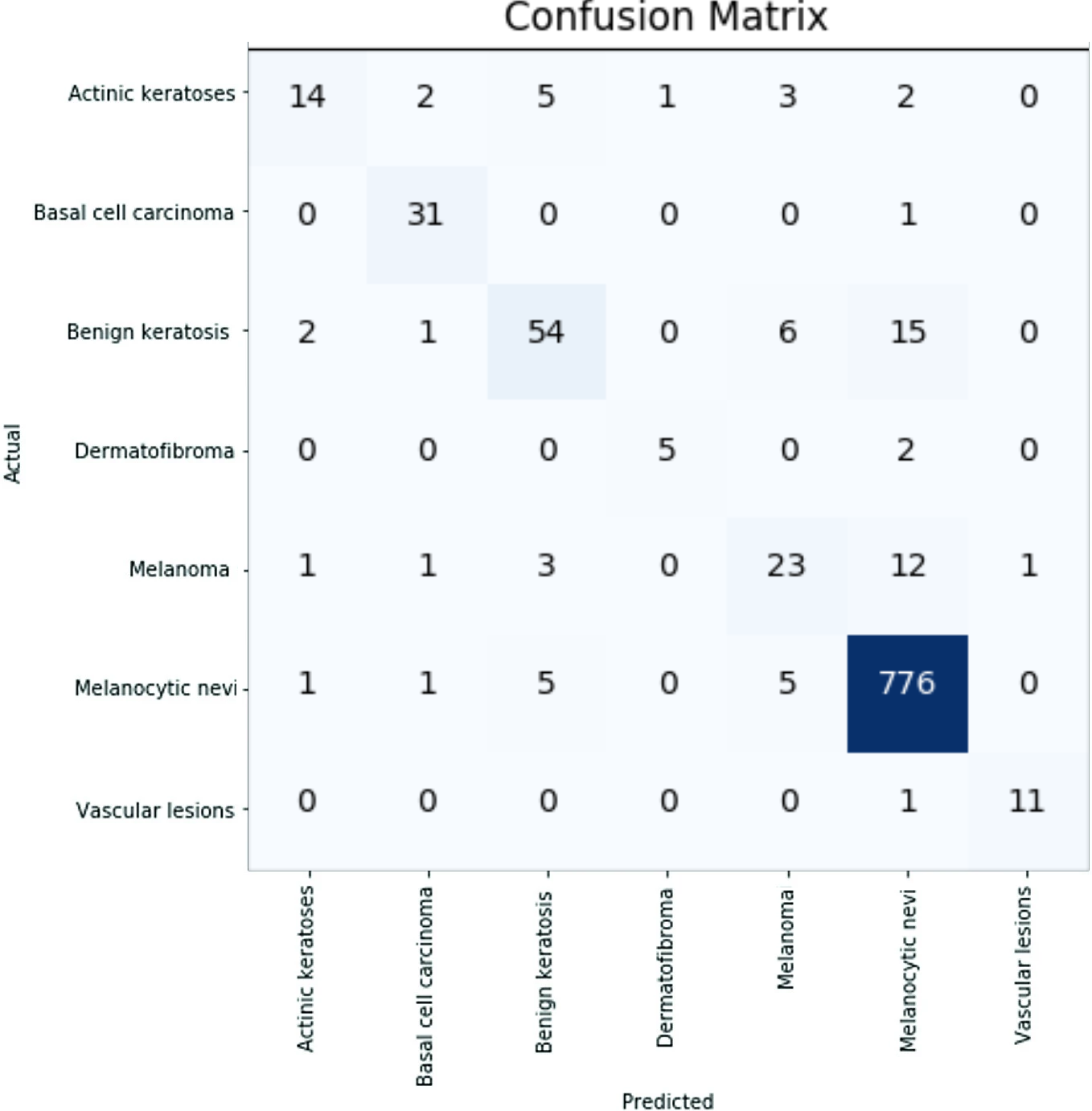}
    \caption{Confusion Matrix ViT\_L16}
    \label{fig:13}
\end{figure}

The confusion matrix for the better performing model, ViT\_L16, can be seen in Figure \ref{fig:13}. Considering the diagonal axis from top left to bottom right, it can be seen that most of the classifications are on this axis. This illustrates the good performance of the model, indicating that the vast majority of skin lesions are correctly classified. However, it can be noticed that a number of melanoma cases are incorrectly classified in the most represented type melanocytic nevi. For the ViT\_L32 model, the corresponding matrix is found in Appendix B: Figure \ref{fig:12}. 

These results of the underlying models are also better compared to those of \citeauthor{korgialas2021vision} \shortcite{korgialas2021vision}, which have an accuracy of 74.73\% (ViT\_B32) and 81.88\% (ViT\_B16). Looking at the recall metrics for the skin cancer type melanoma of the respective models, it can be seen that our two key models deliver better performance. The recall is 42.68\% and 212.53\% higher, respectively, compared to the \citeauthor{korgialas2021vision} \shortcite{korgialas2021vision} models.With both models we also achieve a better accuracy than the best CNN model of \citeauthor{garg2021decision} \shortcite{garg2021decision}. However, only the ViT\_L32 manages to outperform the recall of the CNN model. In addition, a comprehensive ablation study examining different customizations of the ViT\_L16 model architecture and comparing their performance metrics on the basis of accuracy can be found in the Appendix C: Table \ref{tab:ablation_study}.

\begin{figure}[htbp]
    \centering
    \includegraphics[width=\columnwidth]{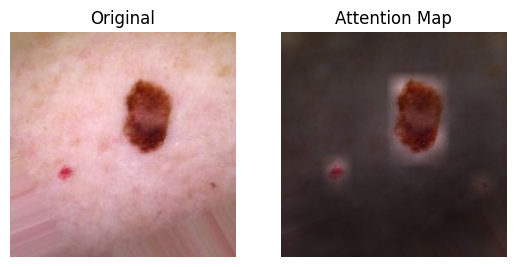}
    \caption{Skin Lesion under Attention Map}
    \label{fig:4}
\end{figure}

A crucial component of the corresponding ViT models is, as already explained, an attention mechanism with 16 heads. The result of this is illustrated using an exemplary skin cancer image in Figure \ref{fig:4}. On the left is the image of the skin lesion and on the right what it looks like after applying the attention mechanism. It can be seen that the mechanism recognizes the lesion and its contours very well. This component is elementary for the superior performance of the ViT models compared to the other models. 

In addition, the results of the recall should be considered in a differentiated manner, as it would be possible to obtain a 'perfect' recall for melanoma. This would require all observations to always be predicted for the melanoma class, as this would result in the number of false negatives being equal to 0. However, this would lead to a significant deterioration, since our main metric is accuracy and we only consider recall in second place, this procedure is not optimal.

\section{Conclusion}

Our study shows that Vision Transformers (ViT) can be more effective than traditional methods like CNNs in classifying skin cancer. The main findings indicate that our ViT models, specifically ViT\_L32 and ViT\_L16, outperform baseline models (DTC and KNN) and previous studies that use smaller ViT configurations or CNNs. ViT\_L16 achieves an accuracy of 92.79\% and a melanoma recall of 56.10\%, demonstrating its potential for accurate skin cancer detection. These findings highlight the effectiveness of ViT models in capturing intricate patterns in dermatoscopic images, owing to their advanced attention mechanisms. 

Nevertheless, this study has limitations. The models are trained on a dataset with uneven representation of skin cancer types, which may have affected their performance on diverse real-world datasets. Furthermore, while the recall for the most lethal skin cancer type in our models is better than most reference models, the metric is slightly below 60\% in both models, meaning that over 40\% with melanoma are classified as something else. Additionally, ViTs, like many deep learning models, can be considered “black-boxes”. It is crucial to understand how these models make decisions, particularly in healthcare applications. The lack of interpretability could be a significant limitation in clinical settings where comprehending the reasoning behind a diagnosis is as important as the diagnosis itself. Moreover, ViT models, particularly larger configurations such as ViT\_L16 and ViT\_L32, necessitate significant computational resources for training and inference. The deployment of these technologies in resource-constrained environments, such as rural clinics or developing countries, may be limited. Lastly, the use of data augmentation techniques to address class imbalance may result in overfitting, particularly if the augmented data does not accurately represent real-world variations.

However, future research could benefit from exploring ways to optimize ViT models for more balanced datasets or using techniques to mitigate class imbalance. In addition, future research could broaden its scope to identify and classify other skin diseases, not just cancer, using ViT models or conduct a long-term study to evaluate the performance and reliability of ViT models over time, and their adaptability to evolving clinical guidelines and practices.
Investigating the integration of ViT models into clinical workflows could pave the way for their practical application in the early detection of skin cancer, potentially reducing the burden on healthcare systems and improving patient outcomes. This study presents the potential for advanced machine learning techniques to be used in medical imaging, which could revolutionize the approach and management of skin cancer diagnosis. The implementation of ViT models as part of clinical decision support systems in hospitals and clinics can assist dermatologists in making more informed and accurate diagnoses. In addition, a practical application of ViT models could be the development of user-friendly mobile applications for preliminary skin lesion analysis, which would provide an accessible tool for early detection.

\bibliographystyle{aaai} 
\bibliography{reference.bib}

\newpage
\FloatBarrier
\appendix
\section{Appendix}

\FloatBarrier
\subsection{Appendix A}

\begin{figure}[htbp]
    \centering
    \includegraphics[width=0.3\textwidth]{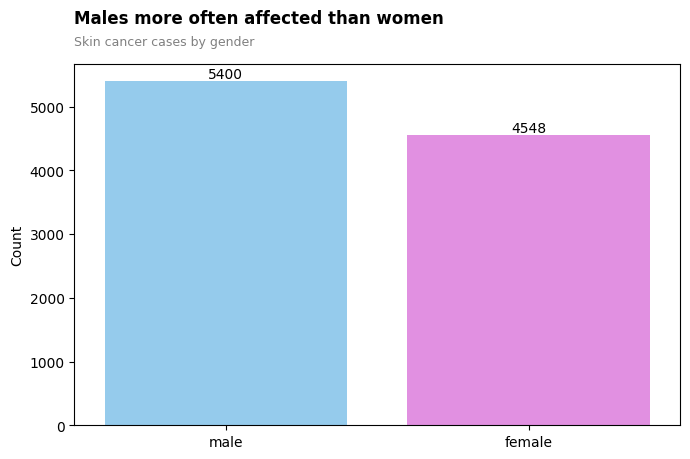}
    \caption{Gender Distribution}
    \label{fig:5}
\end{figure}

\begin{figure}[htbp]
    \centering
    \includegraphics[width=0.3\textwidth]{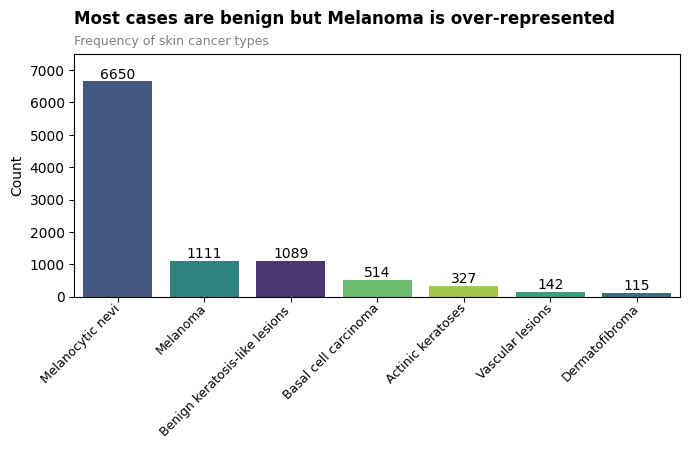}
    \caption{Skin Cancer Types Distribution}
    \label{fig:6}
\end{figure}

\begin{figure}[htbp]
    \centering
    \includegraphics[width=0.3\textwidth]{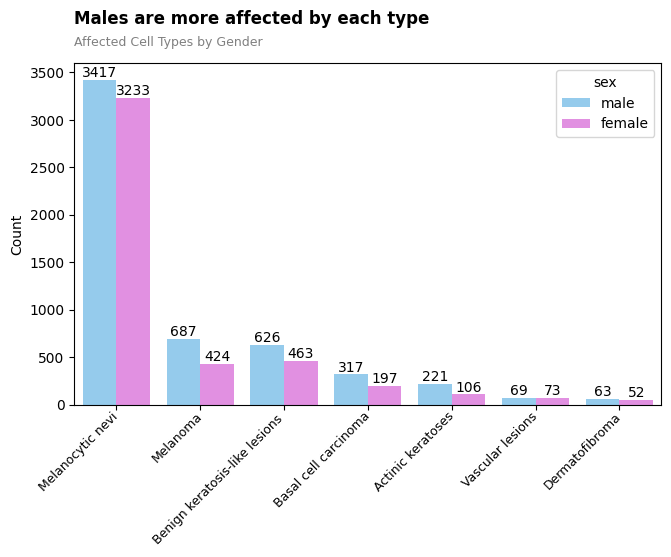}
    \caption{Skin Cancer Types by Gender}
    \label{fig:7}
\end{figure}

\begin{figure}[htbp]
    \centering
    \includegraphics[width=0.3\textwidth]{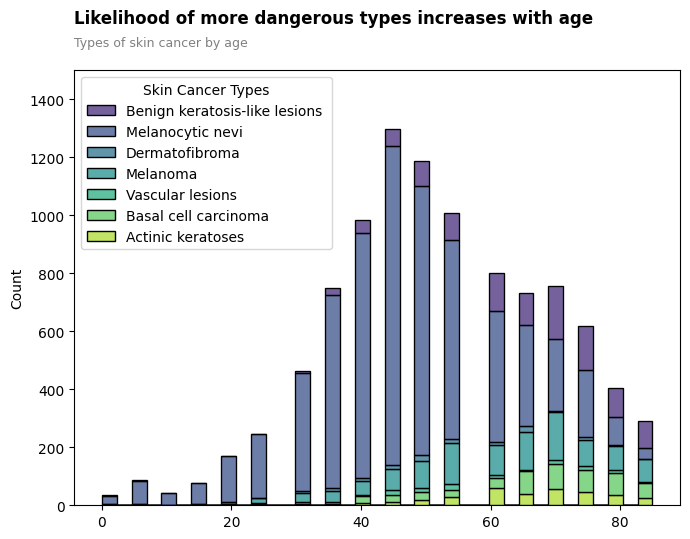}
    \caption{Skin Cancer Types by Age}
    \label{fig:8}
\end{figure}

\begin{figure}[htbp]
    \centering
    \includegraphics[width=0.3\textwidth]{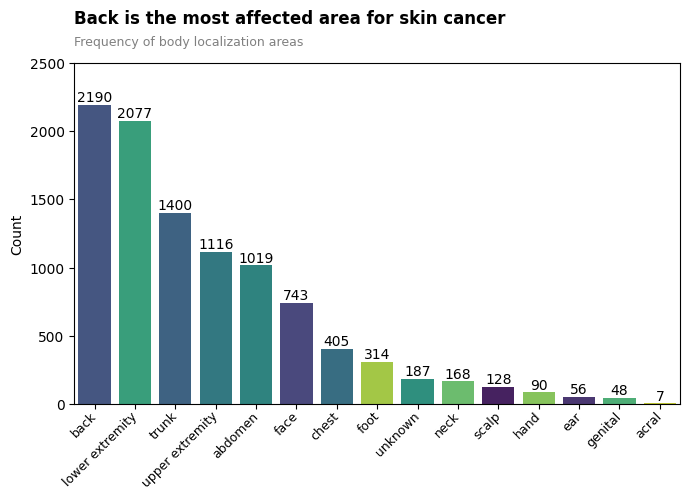}
    \caption{Body Localization Areas Distribution}
    \label{fig:9}
\end{figure}

\begin{figure}[htbp]
    \centering
    \includegraphics[width=0.3\textwidth]{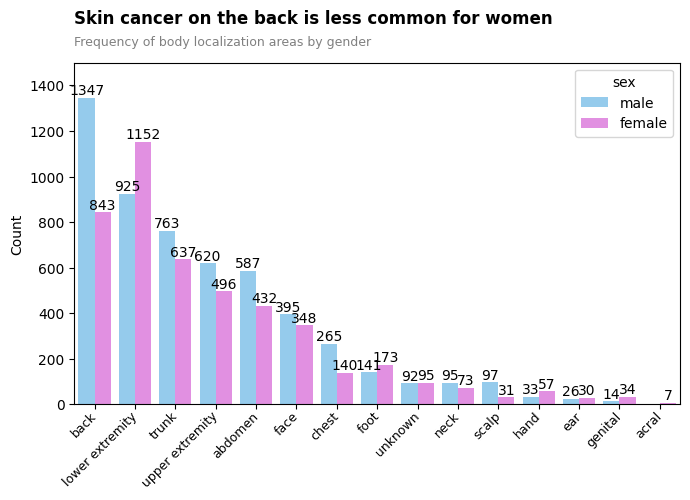}
    \caption{Body Localization Areas by Gender}
    \label{fig:10}
\end{figure}

\begin{figure}[htbp]
    \centering
    \includegraphics[width=0.3\textwidth]{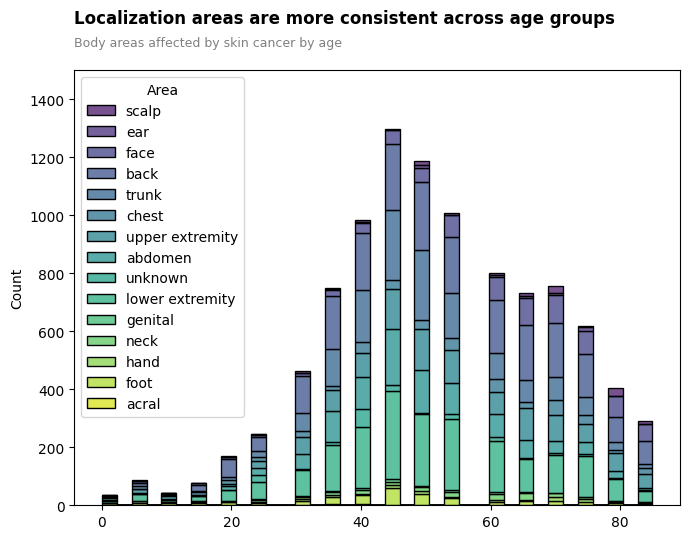}
    \caption{Body Localization Areas by Age}
    \label{fig:11}
\end{figure}

\newpage
\FloatBarrier
\subsection{Appendix B}

\begin{figure}[htbp]
    \centering
    \includegraphics[width=0.3\textwidth]{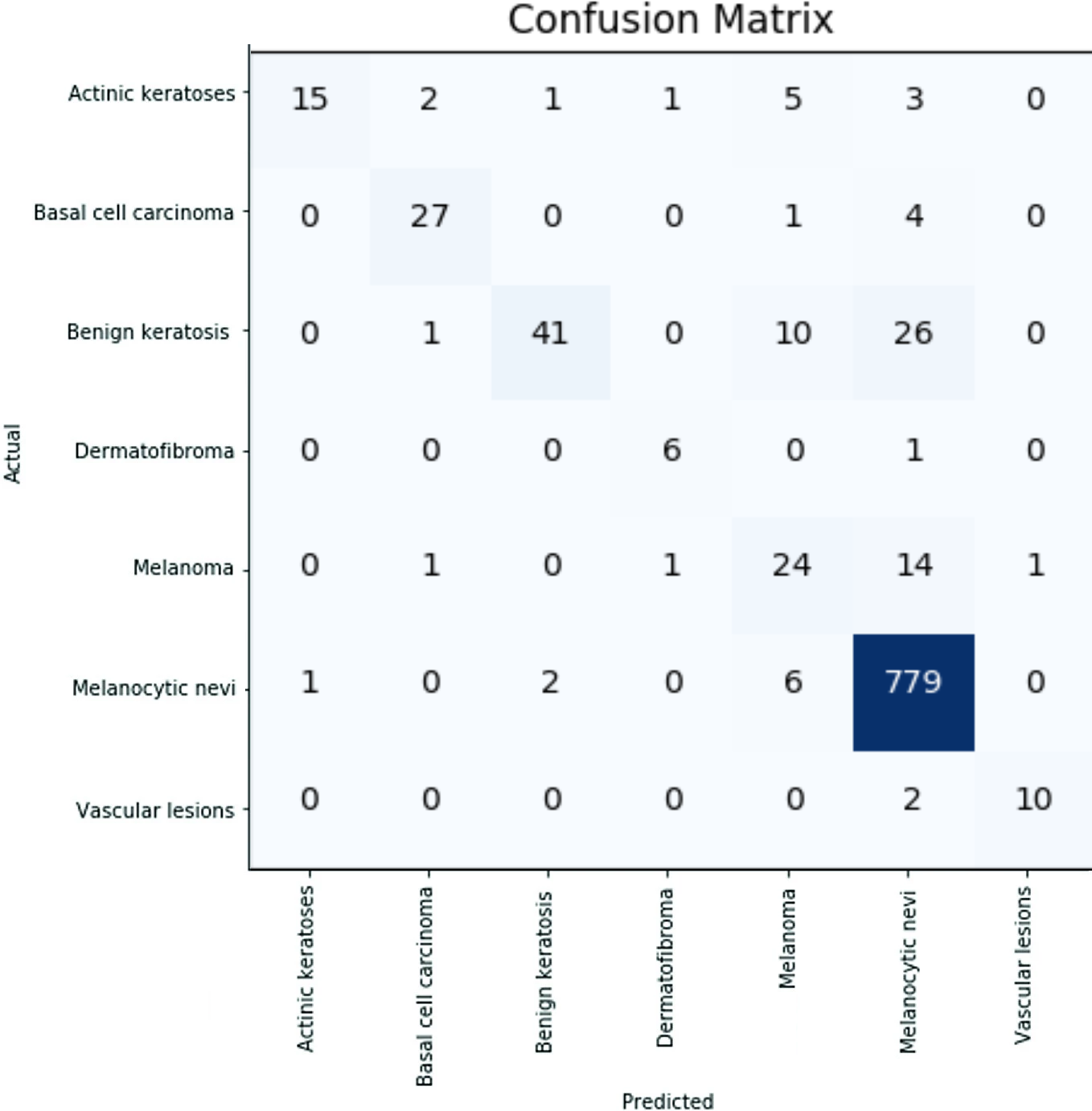}
    \caption{Confusion Matrix ViT\_L32}
    \label{fig:12}
\end{figure}

\subsection{Appendix C}

\begin{table}[htbp]
  \centering
  \caption{Ablation Study Results}
  \label{tab:ablation_study}
  \scriptsize 
  \begin{adjustbox}{width=0.5\textwidth}
  \begin{threeparttable}
  \begin{tabular}{cccccccp{1.2cm}cccc}
    \toprule
    Batch & Epochs & Neurons* & Activation & L2 & Dropout & LR & ReduceLR & Optimizer & Accuracy \\
    Size & & & Function* & Regularization* & Layer & Scheduler** & On Plateau** & & \\
    \midrule
    16 & 10 & 11 & GELU & x & x & \checkmark & x & SGD & 85.58\% \\
    64 & 30 & 11 & GELU & x & x & \checkmark & x & SGD & 83.45\% \\
    16 & 10 & 28 & GELU & x & x & \checkmark & x & SGD & 87.72\% \\
    32 & 20 & 28 & GELU & x & x & \checkmark & x & SGD & 85.89\% \\
    16 & 20 & 28 & RReLU & x & x & \checkmark & x & SGD & 87.21\% \\
    16 & 20 & 28 & ReLU & x & x & \checkmark & x & SGD & 87.72\% \\
    16 & 20 & 28 & ReLU & \checkmark & \checkmark & \checkmark & x & SGD & 86.60\% \\
    16 & 20 & 28 & ReLU & \checkmark & x & \checkmark & x & SGD & 87.51\% \\
    16 & 20 & 28 & ReLU & x & x & x & \checkmark & SGD & 89.95\% \\
    16 & 20 & 28 & ReLU & x & x & x & \checkmark & ADAM & 83.55\% \\
    16 & 20 & 28 & GELU & x & \checkmark & x & \checkmark & ADAM & 85.79\% \\
    16 & 20 & 28 & ReLU & x & \checkmark & x & \checkmark & SGD & 92.79\% \\
    \bottomrule
  \end{tabular}
  \end{threeparttable}
  \end{adjustbox}
  \begin{tablenotes}
  \small
  \item[*] *Part of the dense layer, which comes after the pre-trained ViT, 
  
  flatten and batch normalization layer.
  \item[**] **Part of the callbacks for training the network.
  \end{tablenotes}
\end{table}

\end{document}